\documentclass[10pt,twocolumn,letterpaper]{article}

\usepackage[pagenumbers]{cvpr} 

\usepackage{graphicx}
\usepackage{amsmath}
\usepackage{amssymb}
\usepackage{booktabs}

\usepackage{color}
\usepackage{multirow}
\usepackage[dvipsnames]{xcolor}
\usepackage[switch]{lineno}
\usepackage[accsupp]{axessibility}

\usepackage{colortbl}

\definecolor{lightskyblue}{rgb}{0.94,1.0,1.0}
\definecolor{lightgreen}{rgb}{0.94,1.0,0.94}
\definecolor{whitesmoke}{rgb}{0.92,0.92,0.92}
\definecolor{seashell}{rgb}{1.0,0.96,0.93}

\usepackage[pagebackref,breaklinks,colorlinks]{hyperref}

\usepackage[capitalize]{cleveref}
\crefname{section}{Sec.}{Secs.}
\Crefname{section}{Section}{Sections}
\Crefname{table}{Table}{Tables}
\crefname{table}{Tab.}{Tabs.}

\newcommand{\beginsupplement}{%
        \setcounter{table}{0}
        \renewcommand{\thetable}{S\arabic{table}}%
        \setcounter{figure}{0}
        \renewcommand{\thefigure}{S\arabic{figure}}%
        \renewcommand{\theequation}{S\arabic{equation}}
        \setcounter{section}{0}
        \renewcommand{\thesection}{S\arabic{section}}
 }

\def\cvprPaperID{9143} 
\def\confName{CVPR}
\def\confYear{2023}

\begin{document}

\title{NeuDA: Neural Deformable Anchor for \\ High-Fidelity Implicit Surface Reconstruction}

\author{
Bowen Cai\ \ \ \ \ \
Jinchi Huang\ \ \ \ \ \
Rongfei Jia\ \ \ \ \ \
Chengfei Lv\ \ \ \ \ \
Huan Fu\thanks{Corresponding author.}\ \ \ \ \ \
\\
{Tao Technology Department, Alibaba Group} 
}
\maketitle

\begin{abstract}

   This paper studies implicit surface reconstruction leveraging differentiable ray casting. Previous works such as IDR \cite{yariv2020multiview} and NeuS \cite{wang2021neus} overlook the spatial context in 3D space when predicting and rendering the surface, thereby may fail to capture sharp local topologies such as small holes and structures. To mitigate the limitation, we propose a flexible neural implicit representation leveraging hierarchical voxel grids, namely Neural Deformable Anchor (NeuDA), for high-fidelity surface reconstruction. NeuDA maintains the hierarchical anchor grids where each vertex stores a 3D position (or anchor) instead of the direct embedding (or feature). We optimize the anchor grids such that different local geometry structures can be adaptively encoded. Besides, we dig into the frequency encoding strategies and introduce a simple hierarchical positional encoding method for the hierarchical anchor structure to flexibly exploit the properties of high-frequency and low-frequency geometry and appearance. Experiments on both the DTU \cite{jensen_dtu} and BlendedMVS \cite{yao2020blendedmvs} datasets demonstrate that NeuDA can produce promising mesh surfaces.
   
\end{abstract}

\section{Introduction}
\label{sec:intro}



\begin{figure}[t]
  \centering
   \includegraphics[width=1.0\linewidth]{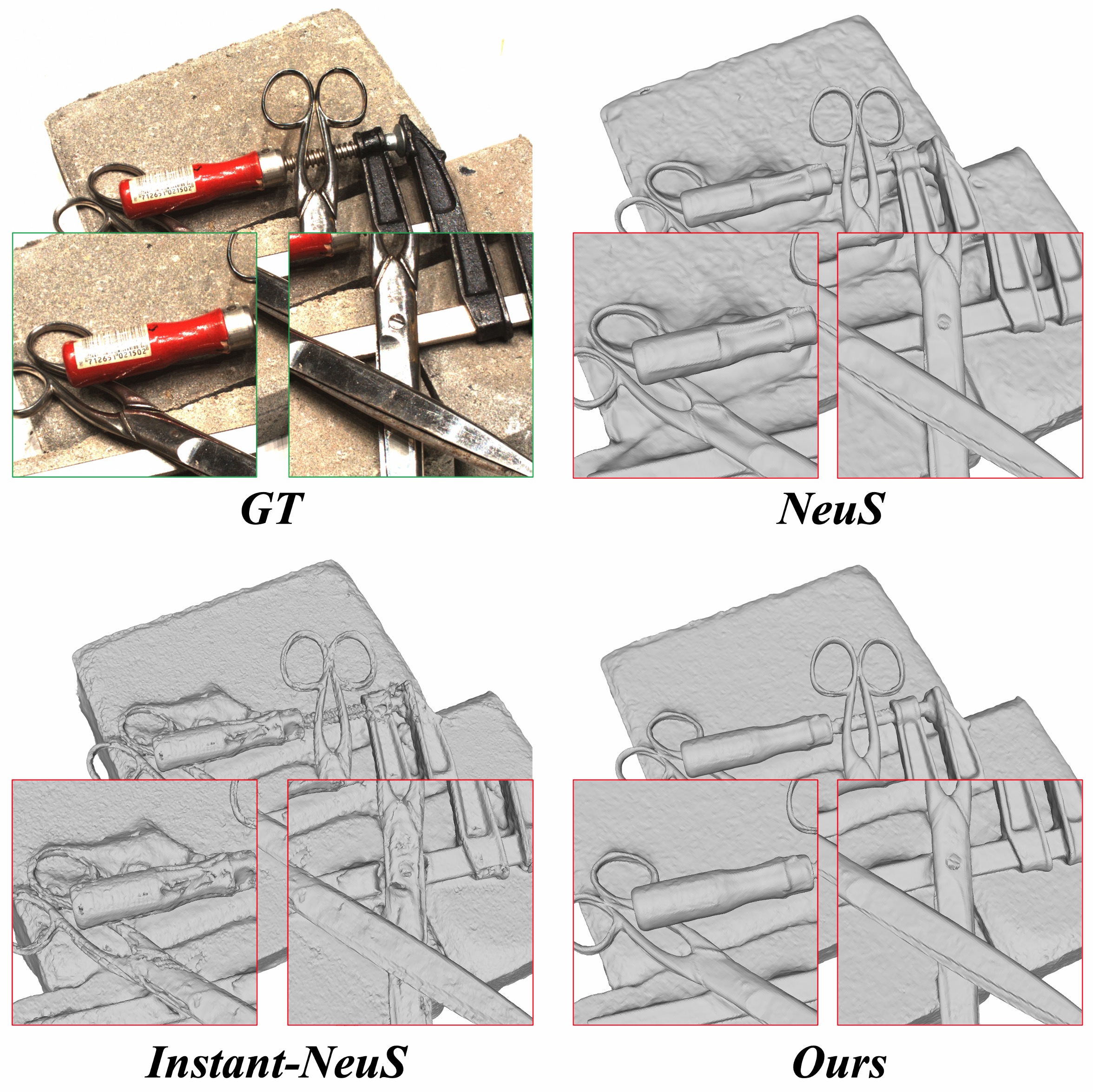}

   \caption{We show the surface reconstruction results produced by NeuDA and the two baseline methods, including NeuS \cite{wang2021neus} and Intsnt-NeuS \cite{wang2021neus,mueller2022instant}. Intsnt-NeuS is the reproduced NeuS leveraging the multi-resolution hash encoding technique \cite{mueller2022instant}. We can see NeuDA can promisingly preserve more surface details. Please refer to Figure~\ref{fig:da-cmp} for more qualitative comparisons.}
   \label{fig:teaser}
\end{figure}

3D surface reconstruction from multi-view images is one of the fundamental problems of the community. Typical Multi-view Stereo (MVS) approaches perform cross-view feature matching, depth fusion, and  surface reconstruction (e.g., Poisson Surface Reconstruction) to obtain triangle meshes \cite{kazhdan2006poisson}. Some methods have exploited the possibility of training end-to-end deep MVS models or employing deep networks to improve the accuracy of sub-tasks of the MVS pipeline. Recent advances show that neural implicit functions are promising to represent scene geometry and appearance \cite{park_deepsdf_2019, wang2021neus, mildenhall2020nerf, niemeyer_differentiable_2020, yariv2021volume, oechsle2021unisurf, yariv2020multiview, mescheder_occupancy_2019, zhang_learning_2021,rosu2022hashsdf,wang2022neus2,ling2022shadowneus, long2022neuraludf}. For example, several works \cite{wang2021neus,yariv2020multiview,wu_voxurf_2022,fu_geo-neus_2022} define the implicit surface as a zero-level set and have captured impressive topologies. Their neural implicit models are trained in a self-supervised manner by rendering faithful 2D appearance of geometry leveraging differentiable rendering. However, the surface prediction and rendering formulations of these approaches have not explored the spatial context in 3D space. As a result, they may struggle to recover fine-grain geometry in some local spaces, such as boundaries, holes, and other small structures (See Fig.~\ref{fig:teaser}).

\begin{figure*}[t]
  \centering
   \includegraphics[width=1.0\linewidth]{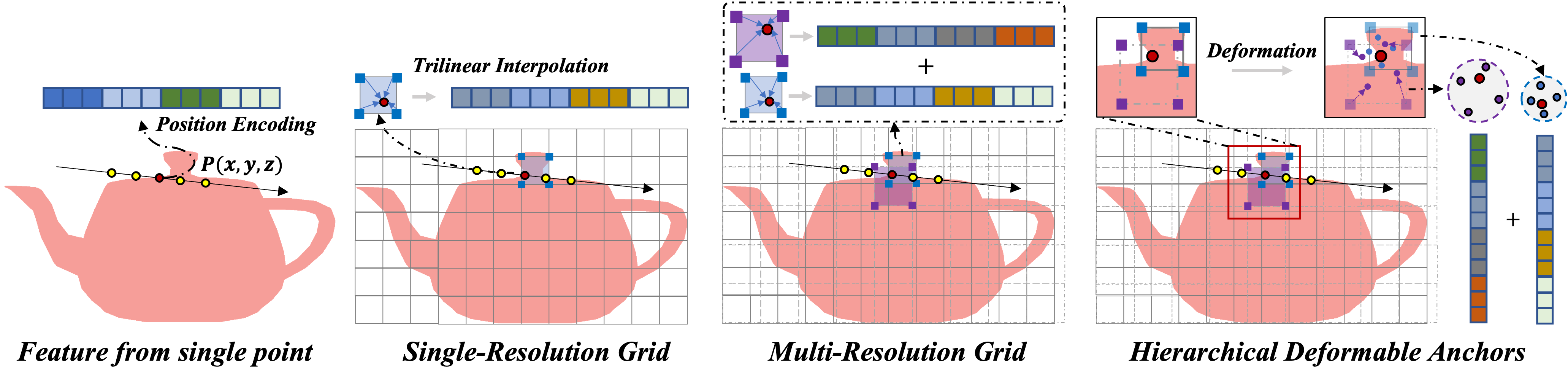}

   \caption{We elaborate on the main differences between the hierarchical deformable anchors representation and some baseline variants. From left to right: (1) Methods such as NeuS \cite{wang2021neus}, volSDF\cite{yariv2021volume}, and UNISUFR\cite{oechsle2021unisurf} sample points along a single ray; (2, 3) Standard voxel grid approaches store a learnable embedding (or) feature at each vertex. Spatial context could be simply handled through the feature aggregation operation. The multi-resolution (or hierarchical) voxel grid representation can further explore different receptive fields; (4) Our method maintains a 3D position (or anchor point) instead of a feature vector at each vertex. We optimize the anchor points such that different geometry structures can be adaptively represented.}
   \label{fig:motivation}
   \vspace{-0.3cm}
\end{figure*}

A straightforward solution is to query scene properties of a sampled 3D point by fusing its nearby features. For example, we can represent scenes as neural voxel fields \cite{sun2022direct,yu_and_fridovichkeil2021plenoxels,liu2020neural,chen2022tensorf,takikawa2021neural} where the embedding (or feature) at each vertex of the voxel encodes the geometry and appearance context. Given a target point, we are able to aggregate the features of the surrounding eight vertices. As the scope of neighboring information is limited by the resolution of grids, multi-level (or hierarchical) voxel grids have been adopted to study different receptive fields \cite{mueller2022instant,wu_voxurf_2022,Yu2022MonoSDF,takikawa2021neural,rosu2022hashsdf,wang2022neus2}. 
These approaches do obtain sharper surface details compared to baselines for most cases, but still cannot capture detailed regions well. A possible reason is that the geometry features held by the voxel grids are uniformly distributed around 3D surfaces, while small structures are with complicated typologies and may need more flexible representations. 
\newline

\noindent \textbf{Contributions:} Motivated by the above analysis, we introduce Neural Deformable Anchor (NeuDA), a new neural implicit representation for high-fidelity surface reconstruction leveraging multi-level voxel grids. Specifically, we store the 3D position, namely the anchor point, instead of the regular embedding (or feature) at each vertex. The input feature for a query point is obtained by directly interpolating the frequency embedding of its eight adjacent anchors. The anchor points are optimized through backpropagation, thus would show flexibility in modeling different fine-grained geometric structures. Moreover, drawing inspiration that high-frequency geometry and texture are likely encoded by the finest grid level, we present a simple yet effective hierarchical positional encoding policy that adopts a higher frequency band to a finer grid level. Experiments on DTU \cite{jensen_dtu} and BlendedMVS \cite{yao2020blendedmvs} shows that NeuDA is superior in recovering high-quality geometry with fine-grains details in comparison with baselines and SOTA methods. It's worth mentioning that NeuDA employs a shallower MLP (4 \emph{vs.} 8 for NeuS and volSDF) to achieve better surface reconstruction performance due to the promising scene representation capability of the hierarchical deformable anchor structure.

\section{Related Work}
\label{sec:relatedwork}

\noindent \textbf{Neural Implicit Surface Reconstruction}
Recently, neural surface reconstruction has emerged as a promising alternative to traditional 3D reconstruction methods due to its high reconstruction quality and its potential to recover fine details. NeRF \cite{mildenhall2020nerf} proposes a new avenue combining neural implicit representation with volume rendering to achieve high-quality rendering results. The surface extracted from NeRF often contains conspicuous noise; thus, its recovered geometry is far from satisfactory. To obtain an accurate scene surface, DVR \cite{niemeyer_differentiable_2020}, IDR \cite{yariv2020multiview}, and NLR \cite{kellnhofer2021neural} have been proposed to use accurate object masks to promote reconstruction quality. Furthermore, NeuS \cite{wang2021neus}, UNISURF \cite{oechsle2021unisurf}, and volSDF \cite{yariv2021volume} learn an implicit surface via volume rendering without the need for masks and shrink the sample region of volume rendering to refine the reconstruction quality. 
Nevertheless, the above approaches extract geometry features from a single point along a casting ray, which may hinder the neighboring information sharing across sampled points around the surface. 
The quality of the reconstructed surface depends heavily on the capacity of the MLP network to induce spatial relationships between neighboring points. Thereby, NeuS \cite{wang2021neus}, IDR \cite{yariv2020multiview}, and VolSDF \cite{yariv2021volume} adopt deep MLP network and still struggle with fitting smooth surfaces and details. 
It is worth mentioning that the Mip-NeRF \cite{barron2021mip} brings the neighboring information into the rendering procedure by tracing an anti-aliased conical frustum instead of a ray through each pixel. But it is difficult to apply this integrated positional encoding to surface reconstruction since this encoding relies on the radius of the casting cone.
\newline

\noindent \textbf{Neural Explicit Representation}
The neural explicit representation that integrates traditional 3D representation methods, e.g. voxels \cite{liu2020neural,sun2022direct,yu2021plenoctrees} and point clouds \cite{xu2022point}, has made great breakthroughs in recent years. This explicit representation makes it easier to inject the neighborhood information into the geometry feature during model optimization. DVGO \cite{sun2022direct} and Plenoxels \cite{yu_and_fridovichkeil2021plenoxels} represent the scene as a voxel grid, and compute the opacity and color of each sampled point via trilinear interpolation of the neighboring voxels. The Voxurf \cite{wu_voxurf_2022} further extends this single-level voxel feature to a hierarchical geometry feature by concatenating the neighboring feature stored voxel grid from different levels. The Instant-NGP \cite{mueller2022instant} and MonoSDF \cite{Yu2022MonoSDF} use multiresolution hash encoding to achieve fast convergence and capture high-frequency and local details, but they might suffer from hash collision due to its compact representation. Both of these methods leverage a multi-level grid scheme to enlarge the receptive field of the voxel grid and encourage more information sharing among neighboring voxels. Although the voxel-based methods have further improved the details of surface geometry, they may be suboptimal in that the geometry features held by the voxel grids are uniformly distributed around 3D surfaces, while small structures are with complicated typologies and may need more flexible representation.


Point-based methods \cite{xu2022point,boulch2022poco,liang_spidr_2022} bypass this problem, since the point clouds, initially estimated from COLMAP \cite{schonberger_structure--motion_2016}, are naturally distributed on the 3D surface with complicated structures. Point-NeRF \cite{xu2022point} proposes to model point-based radiance field, which uses an MLP network to aggregate the neural points in its neighborhood to regress the volume density and view-dependent radiance at that location. However, the point-based methods are also limited in practical application, since their reconstruction performance depends on the initially estimated point clouds that often have holes and outliers.


\begin{figure*}[t]
  \centering
   \includegraphics[width=0.95\linewidth]{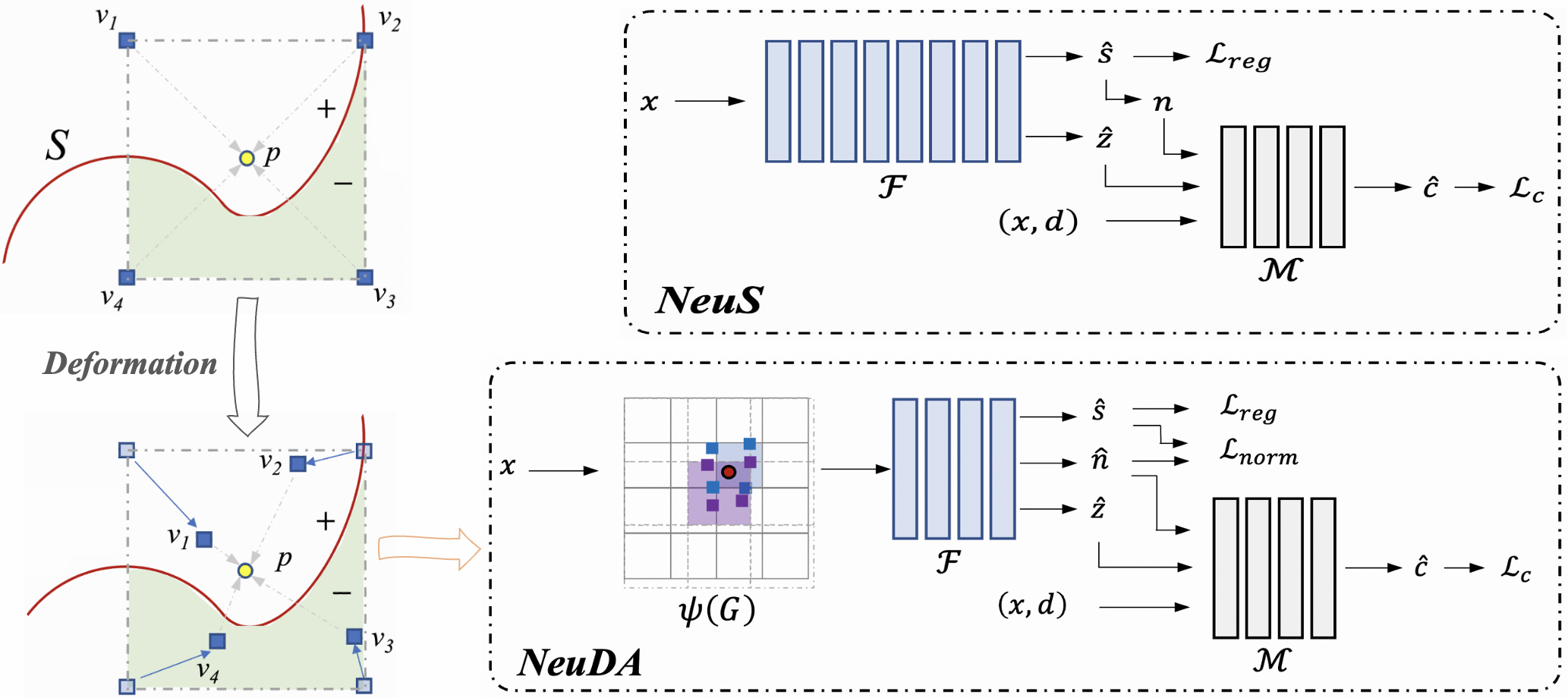}

   \caption{\textbf{\emph{Left:}} An intuitive explanation of the deformable anchors (DA) approach. The anchor points ($v$) surrounding the target point $p$ would more flexibly encode the local geometry as the model converges. \textbf{\emph{Right:}} Overall, NeuDA shares a similar architecture with NeuS, except for the DA part $\psi(G)$ (Sec.~\ref{deformable_anchor}), the involved HPE policy (Sec.~\ref{hierarchical_pe}), and a normal regularization term $\mathcal{L}_{norm}$ \cite{verbin2022ref}. Another difference is that NeuDA is capable of a shallower SDF network $\mathcal{F}$ (4 \emph{vs.} 8 for NeuS).
   }
   \label{fig:deformable_anchor}
\end{figure*}

\section{Method}

Our primary goal is to flexibly exploit spatial context around the object surfaces to recover more fine-grained typologies, as a result, boost the reconstruction quality. This section begins with a brief review of NeuS \cite{wang2021neus}, which is our main baseline, in Sec. \ref{preliminaries}. Then, we explain the deformable anchor technique in Sec. \ref{deformable_anchor}, and present the hierarchical position encoding policy in Sec. \ref{hierarchical_pe}. Finally, we present the objectives and some optimization details of NeuDA in Sec. \ref{opt_details}.

\subsection{Preliminaries: NeuS}
\label{preliminaries}
NeuS \cite{wang2021neus} is one of the promising neural implicit surface reconstruction approaches that smartly takes advantage of both the IDR \cite{yariv2020multiview} and NeRF \cite{mildenhall2020nerf} formulations. It represents the geometry as the zero-level set of signed distance function (SDF) $\mathcal{S} = \left \{x \in \mathbb{R}^3 | f(x)=0  \right \}$, and alleviates the discernible bias issue of standard volume rendering to learn a better SDF representation. The signed distance function is parameterized with a 8-layer MLP $\mathcal{F}(x;\theta) = \left ( f(x;\theta ), z(x;\theta ) \right ) \in \mathbb{R} \times \mathbb{R}^{256}$, where $z(x;\theta )$ is the learned geometric property from the 3D point (or position) $x \in \mathbb{R}^3$. And a 4-layer MLP $\mathcal{M}(x, d, n, \hat{z};\gamma ) \in \mathbb{R}^3$ is adopted to approximate the color from the factors such as the view direction $d$, normal $n$, and geometric feature $\hat{z} = z(x;\theta)$.


To render a pixel, a ray $\left \{ p(t) = o+td | t>0 \right \}$ is emitted from the camera center $o$ along the direction $d$ passing through this pixel. The rendered color $\hat{C}$ for this pixel is accumulated along the ray with $N$ discrete sampled points:

\begin{equation}
    \begin{aligned}
      \hat{C} = \sum_{i=1}^{N}T_i\alpha_i c_i, \quad T_i=\prod_{j=1}^{i-1}(1-\alpha_j) 
    \end{aligned}
\end{equation}
where $T_i$ denotes accumulated transmittance. $\alpha_i$ represents discrete opacity. To ensure unbiased surface reconstruction in the first-order approximation of SDF, NeuS defines the opacity as follows:
\begin{equation}
    \begin{aligned}
      \alpha_i = \text{max}\left ( \frac{\Phi_s(f(p(t_i)))- \Phi_s(f(p(t_{i+1})))}{\Phi_s(f(p(t_i)))}, 0 \right ) 
    \end{aligned}
\end{equation}
Here, $\Phi_s(x)$ is constructed based on the probability density function, defined by $\Phi_s(x) =(1+e^{-sx})^{-1}$. The $s$ value is a trainable parameter, and $1/s$ approaches to zero as the optimization converges.


\subsection{Deformable Anchors (DA)}
\label{deformable_anchor}

Our motivation for proposing the deformable anchors technique is to improve the flexibility of the voxel grid representation such that the spatial context in 3D space can be better exploited. Overall, we assign a 3D position (or anchor point) rather than a feature vector at each vertex, as depicted in Figure \ref{fig:deformable_anchor}. We optimize the anchor points such that they can adaptively move from the corners to the vicinity of the abrupt geometry-changing area as training convergences. In the following, we will take a sample point $p \in \mathbb{R}^3$ along a specific ray as an example to explain the DA representation in a single-level grid. We use 8-nearest neighbor anchors to characterize the sample point.

Specifically, we first normalize the input coordinate of $p$ to the grid's scale. The normalized coordinate is denoted as $x$. Then, we map the sample point to a voxel via $\mathcal{V} = \left \{ v | \left \lfloor x*N \right \rfloor  <= v < \left \lceil x*N \right \rceil  \right \} $. Here, $N$ represents the size of the grid, and the anchor points are stored at the eight vertices of a voxel.
Finally, we can conveniently obtain the input feature, which would be fed into $\mathcal{F}$ in Figure~\ref{fig:deformable_anchor}, by interpolating the frequency embedding of these eight adjacent anchors:
\begin{equation}
    \label{eqn:interpolation}
    \begin{aligned}
      \phi(p, \psi(G)) &= \sum_{v \in \mathcal{V}} w(p_v) \cdot \gamma (p_v + \bigtriangleup{p_v}), \\
      \psi(G) &= \left \{ p_v, \bigtriangleup{p_v}| v \in {G} \right \}.
    \end{aligned}
\end{equation}
where $G$ denotes the anchor grid, $\psi(G)$ is a set of deformable anchors that, in the beginning, are uniformly distributed at voxel vertices, and $\gamma(p_v + \bigtriangleup{p_v} )$ is a frequency encoding function. We use cosine similarity as weight $w(p_n)$ to measure the contributions of different anchors to the sampled point:
\begin{equation}
    \begin{aligned}
       w(p_n) = \frac{\hat{w}(p_n)}{\sum_n \hat{w}(p_n)}, \quad \hat{w}(p_n) =   \frac{p\cdot p_n}{\left \| p  \right \|\left \| p_n \right \|  }.
    \end{aligned}
\end{equation}

Given the definition of deformable anchors above, we can approximate the SDF function $f(x;\theta )$, normal $\hat{n}(x;\theta)$, and geometric feature $z(x;\theta )$ of the target object as follows:
\begin{equation}
    \begin{aligned}
       \mathcal{F}(x; \theta) = \mathcal{F}\left ( \phi \left (p, \psi(G) \right ) ;  \theta \right ) \\
        = \left ( f(x;\theta ), \hat{n}(x;\theta ), z(x;\theta ) \right ).
    \end{aligned}
\end{equation}

\subsection{Hierarchical Positional Encoding}
\label{hierarchical_pe}
We employ multi-level (or hierarchical) anchor grids to consider different receptive fields. Following previous works \cite{wang2021neus,yariv2021volume}, we utilize positional encoding to better capture high-frequency details. But as we have several levels of anchor grid (8 levels in this paper), applying the standard positional encoding function \cite{mildenhall2020nerf} to each level followed by a concatenation operation would produce a large-dimension embedding. We argue that different anchor grid levels could have their own responsibilities for handling global structures or capturing detailed geometry variations.

Mathematically, given an anchor point $p_l \in \mathbb{R}^3$ in a specific level $l$, the frequency encoding function $\gamma(p_l)$ follows the below formulation:
\begin{equation}
\label{eqn:HPE-1}
    \begin{aligned}
        \gamma(p_l) &= \left ( \text{sin}(2^l \pi p_l), \text{cos}(2^l \pi p_l) \right ). \\
    \end{aligned}
\end{equation}
The frequency function $\gamma(p_l)$ is applied to the three coordinate values in $p_l$ individually. Then, the interpolation operation in Eqn.~\ref{eqn:interpolation} would return a small 6-dimension embedding $\phi(\hat{p}_l)$ for each anchor grid level. Finally, we concatenate multi-level embedding vectors to obtain the hierarchical positional encoding:
\begin{equation}
\label{eqn:HPE-2}
    \begin{aligned}
       \mathcal{H}(p) &= (\phi(\hat{p}_0), \phi(\hat{p}_1), ..., \phi(\hat{p}_{L-1})),
    \end{aligned}
\end{equation}
where $L$ is the total grid level which is set to 8 in our experiments if not specified. The encoded hierarchical feature will be fed into the SDF network to predict the signed distance $f\left ( \mathcal{H}(p); \theta \right )$.

\setlength\tabcolsep{3.7pt}
\begin{table*}[th!]
\centering
\begin{tabular}{ c | c c c c c c c c c c c c c c c | c }
\toprule
ScanID & 24 & 37 & 40 & 55 & 63 & 65 & 69 & 83 & 97 & 105 & 106 & 110 & 114 & 118 & 122 & Mean \\
\midrule\midrule
\multicolumn{17}{c}{ w/ mask } \\
\midrule
IDR \cite{yariv2020multiview} & 1.63 & 1.87 & 0.63 & 0.48 & 1.04 & 0.79 & 0.77 & 1.33 & 1.16 & 0.76 & 0.67 & 0.9 & 0.42 & 0.51 & 0.53 & 0.90 \\
Voxurf \cite{wu_voxurf_2022} & 0.65 & 0.74 & 0.39 & 0.35 & 0.96 & 0.64 & 0.85 & 1.58 & 1.01 & 0.68 & 0.6 & 1.11 & 0.37 & 0.45 & 0.47 & 0.72 \\
NeuS \cite{wang2021neus} & 0.83 & 0.98 & 0.56 & 0.37 & 1.13 & 0.59 & 0.60 & 1.45 & 0.95 & 0.78 & 0.52 & 1.43 & 0.36 & 0.45 & 0.45 & 0.77 \\
Instant-NeuS \cite{mueller2022instant,wang2021neus} & 0.60 & 1.03 & 0.39 & 0.35 & 1.38 & 0.64 & 0.69 & 1.45 & 1.48 & 0.82 & 0.53 & 1.15 & 0.38 & 0.60 & 0.48 & 0.80 \\
\midrule
NeuDA & 0.51 & 0.76 & 0.39 & 0.37 & 1.08 & 0.56 & 0.57 & 1.37 & 1.13 & 0.79 & 0.50 & 0.80 & 0.34 & 0.42 & 0.46 & \textbf{0.67} \\
\midrule\midrule
\multicolumn{17}{c}{ w/o mask } \\
\midrule
$\text{colmap}_0$ \cite{schonberger_structure--motion_2016} & 0.81 & 2.05 & 0.73 & 1.22 & 1.79 & 1.58 & 1.02 & 3.05 & 1.4 & 2.05 & 1.00 & 1.32 & 0.49 & 0.78 & 1.17 & 1.36 \\
UNISURF \cite{oechsle2021unisurf} & 1.32 & 1.36 & 1.72 & 0.44 & 1.35 & 0.79 & 0.80 & 1.49 & 1.37 & 0.89 & 0.59 & 1.47 & 0.46 & 0.59 & 0.62 & 1.02 \\
volSDF \cite{yariv2021volume} & 1.14 & 1.26 & 0.81 & 0.49 & 1.25 & 0.70 & 0.72 & 1.29 & 1.18 & 0.70 & 0.66 & 1.08 & 0.42 & 0.61 & 0.55 & 0.86 \\
NeuralWarp \cite{darmon2022improving} & 0.49 & 0.71 & 0.38 & 0.38 & 0.79 & 0.81 & 0.82 & 1.2 & 1.06 & 0.68 & 0.66 & 0.74 & 0.41 & 0.63 & 0.51 & 0.68 \\
Voxurf \cite{wu_voxurf_2022} & 0.71 & 0.78 & 0.43 & 0.35 & 1.03 & 0.76 & 0.74 & 1.49 & 1.04 & 0.74 & 0.51 & 1.12 & 0.41 & 0.55 & 0.45 & 0.74 \\
HF-NeuS \cite{wang_hf-neus_2022} & 0.76 & 1.32 & 0.70 & 0.39 & 1.06 & 0.63 & 0.63 & 1.15 & 1.12 & 0.80 & 0.52 & 1.22 & 0.33 & 0.49 & 0.50 & 0.77 \\
NeuS \cite{wang2021neus} & 1.00 & 1.37 & 0.93 & 0.43 & 1.10 & 0.65 & 0.57 & 1.48 & 1.09 & 0.83 & 0.52 & 1.20 & 0.35 & 0.49 & 0.54 & 0.84 \\
Instant-NeuS \cite{mueller2022instant,wang2021neus} & 0.59 & 0.91 & 0.97 & 0.35 & 1.21 & 0.64 & 0.84 & 1.31 & 1.44 & 0.79 & 0.62 & 1.09 & 0.53 & 0.80 & 0.50 & 0.84 \\
\midrule
$\text{NeuDA}^\dagger$ & 0.53 & 0.74 & 0.41 & 0.36 & 0.93 & 0.64 & 0.58 & 1.33 & 1.08 & 0.75 & 0.48 & 1.03 & 0.34 & 0.41 & 0.42 & 0.67 \\
NeuDA & 0.47 & 0.71 & 0.42 & 0.36 & 0.88 & 0.56 & 0.56 & 1.43 & 1.04 & 0.81 & 0.51 & 0.78 & 0.32 & 0.41 & 0.45 & \textbf{0.65} \\
\bottomrule
\end{tabular} 
\caption{\textbf{Quantitative Comparisons on DTU.} We compare the proposed method to the main baselines, \emph{i.e.}, NeuS and Instant-NeuS, and other SOTA methods using their released codes following their best configurations. $\text{NeuDA}^\dagger$ means we remove the normal regularization term in Eqn.~\ref{eqn:normal}. Overall, NeuDA yields remarkable improvements upon baselines, and achieves the best performance on the DTU dataset under both the w/ mask and w/o mask settings.}
\label{tab:performance-dtu}
\end{table*}

\subsection{Objectives}
\label{opt_details}
We minimize the mean absolute errors between the rendered and ground-truth pixel colors as the indirect supervision for the SDF prediction function:
\begin{equation}
    \begin{aligned}
      \mathcal{L}_{c} = \frac{1}{\mathcal{R}} \sum_{r \in \mathcal{R}} {\left \| C(r) - \hat{C}(r) \right \|  },
    \end{aligned}
\end{equation}
where $r$ is a specific ray in the volume rendering formulation \cite{mildenhall2020nerf}, and $C(r)$ is the corresponded ground truth color. 

We adopt an Eikonal term \cite{gropp_implicit_2020} on the sample points to regularize SDF of $f \left (  \mathcal{H}(p) ;  \theta \right )$ as previously:
\begin{equation}
    \begin{aligned}
      \mathcal{L}_{reg} = \frac{1}{\mathcal{R}N}\sum_{r,i}(\left \| \nabla f(\mathcal{H}(p_{r,i}) )  \right \|_2  - 1 )^2,
    \end{aligned}
\end{equation}
where $N$ is the number of sample points along each ray, and $i$ denotes a specific sample point. 

We optimize the BCE loss term if the ground truth masks are incorporated into the training process:
\begin{equation}
    \begin{aligned}
      \mathcal{L}_{mask} = \text{BCE}( m_r, \sum_{i}^{n}T_{r,i}\alpha_{r,i} ),
    \end{aligned}
\end{equation}
where $m_r$ is the mask label of ray $r$.

In addition to above terms, we study a normal regularization loss \cite{verbin2022ref} in NeuDA. Specifically, we auxiliarily predict a normal vector $\hat{n}_{r,i}$ for each spatial point from $\mathcal{F}$ in Figure~\ref{fig:deformable_anchor}. Hereby, we can tie the gradient of SDF to the predicted normal via:
\begin{equation}
\label{eqn:normal}
    \begin{aligned}
      \mathcal{L}_{norm} = \sum_{r,i} T_{r,i} \alpha_{r,i} \left \| \nabla f(\mathcal{H}(p_{r,i}) ) - \hat{n}_{r,i} \right \| 
    \end{aligned}
\end{equation}
As reported in Table~\ref{tab:performance-dtu} ($\text{NeuDA}^\dagger$ \emph{vs.} NeuDA), under the w/o mask setting, $\mathcal{L}_{norm}$ can slightly boost the mean CD score from 0.67 to 0.65.

Finally, the full objective is formulated as:
\begin{equation}
    \begin{aligned}
      \mathcal{L} = \mathcal{L}_{c} + \lambda_{eik}\mathcal{L}_{reg} + \lambda_{norm} \mathcal{L}_{norm} + \lambda_{mask} \mathcal{L}_{mask}.
    \end{aligned}
\end{equation}
In our experiments, the trade-off parameters $\lambda_{eik}$, $\lambda_{norm}$, and $\lambda_{mask}$ are set to 0.1, $3\times 10^{-5}$, and 0.1, respectively.

\begin{figure*}[th!]
  \centering
   \includegraphics[width=0.98\linewidth]{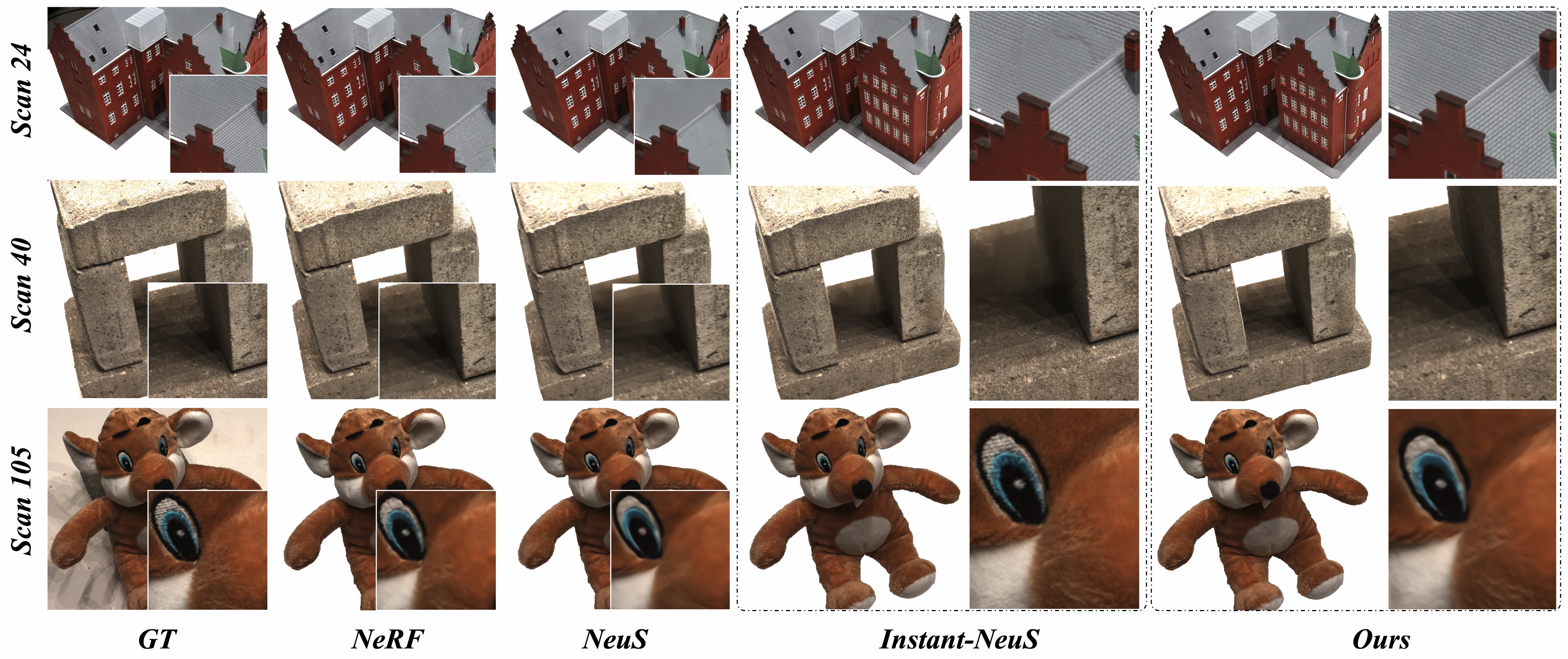}
   \caption{We show several rendered images here. NeuDA produces competitive renderings compared to Instant-NeuS \cite{wang2021neus,mueller2022instant}. Please refer to Table~\ref{tab:psnr} for case-by-case metrics.}
   \label{fig:sota-cmp}
\end{figure*}

\setlength\tabcolsep{2.8pt}
\begin{table*}[th!]
\centering
\small
\begin{tabular}{ c | c c c c c c c c c c c c c c c | c }
\toprule
ScanID & 24 & 37 & 40 & 55 & 63 & 65 & 69 & 83 & 97 & 105 & 106 & 110 & 114 & 118 & 122 & Mean \\
\midrule\midrule
NeRF \cite{mildenhall2020nerf} & 26.24 & 25.74 & 26.79 & 27.57 & 31.96 & 31.50 & 29.58 & 32.78 & 28.35 & 32.08 & 33.49 & 31.54 & 31.0 & 35.59 & 35.51 & 30.65 \\
VolSDF \cite{yariv2021volume} & 26.28 & 25.61 & 26.55 & 26.76 & 31.57 & 31.50 & 29.38 & 33.23 & 28.03 & 32.13 & 33.16 & 31.49 & 30.33 & 34.90 & 34.75 & 30.38 \\
NeuS \cite{wang2021neus} & 25.87 & 23.06 & 26.16 & 25.42 & 29.62 & 28.20 & 28.46 & 32.38 & 26.68 & 30.54 & 28.64 & 28.63 & 28.11 & 31.97 & 34.07 & 28.52 \\
Instant-NeuS \cite{mueller2022instant,wang2021neus} & 30.32 & 25.40 & 30.18 & 32.08 & 31.75 & 29.46 & 29.53 & 33.59 & 27.97 & 32.65 & 32.70 & 31.57 & 30.25 & 35.61 & 36.84 & \textbf{31.33} \\
NeuDA & 29.38 & 24.53 & 27.42 & 27.06 & 33.17 & 33.42 & 29.08 & 34.92 & 28.69 & 32.51 & 33.03 & 29.53 & 29.09 & 36.06 & 35.68 & 30.90   \\
\bottomrule
\end{tabular} 
\caption{We follow the VolSDF setting \cite{yariv2021volume} to report the train PSNR values of our renderings compared to previous works. The multi-resolution hash encoding approach \cite{mueller2022instant} might be a better option than NeuDA for producing high-quality renderings. It's worth mentioning that measuring the rendering quality is just a supplementary experiment, as our main goal is to reconstruct better surfaces.}
\label{tab:psnr}
\end{table*}

\section{Experiments}

This section conducts experiments to validate our NeuDA method for surface reconstruction. First, we take a brief introduction to the studied DTU \cite{jensen_dtu} and BlendedMVS \cite{yao2020blendedmvs} datasets in Sec. \ref{datasets}. Then, we make quantitative and qualitative comparisons with baselines (\emph{e.g.} Instant-NeuS \cite{wang2021neus,mueller2022instant}) and other SOTA neural surface reconstruction approaches in Sec.\ref{benchmark_cmp}. Finally, we present various ablations to discuss NeuDA in Sec. \ref{ablation}. We refer to the supplementary for more experimental results and discussions.

\subsection{Datasets}
\label{datasets}
\noindent \textbf{DTU.} The DTU dataset \cite{jensen_dtu} consists of different static scenes with a wide variety of materials, appearance, and geometry, where each scene contains 49 or 64 images with the resolution of 1600 x 1200. We use the same 15 scenes as IDR \cite{liang_spidr_2022} to evaluate our approach. Experiments are conducted to investigate both the with (w/) and without (w/o) foreground mask settings. As DTU provides the ground truth point clouds, we measure the recovered surfaces through the commonly studied Chamfer Distance (CD) for quantitative comparisons.
\newline

\noindent \textbf{BlendedMVS.} The BlendedMVS dataset \cite{yao2020blendedmvs} consists of a variety of complex scenes, where each scene provides 31 to 143 multi-view images with the image size of 768 $\times$576. We use the same 7 scenes as NeuS \cite{wang2021neus} to validate our method. We only present qualitative comparisons on this dataset, because the ground truth point clouds are not available.

\subsection{Benchmark Comparisons}
\label{benchmark_cmp}

We mainly take NeuS \cite{wang2021neus} and Instant-NeuS \cite{wang2021neus,mueller2022instant} as our baselines. Here, Instant-NeuS is our reproduced NeuS leveraging the multi-resolution hash encoding technique \cite{mueller2022instant}. We also report the scores of some other great implicit surface reconstruction approaches such as UNISURF \cite{oechsle2021unisurf}, volSDF \cite{yariv2021volume}, and NeuralWarp \cite{darmon2022improving}. Unlike other approaches, NeuDA and Instant-NeuS parameterize the SDF function with a slightly shallower MLP (4 \emph{vs.} 8 for NeuS).

The quantitative scores are reported in Table~\ref{tab:performance-dtu}. NeuDA improves the baselines by significant margins under both the w/ mask ($+0.10\sim0.13$) and w/o mask ($+0.19$) settings, and achieves the best mean CD score compared to previous SOTA approaches. Specifically, NeuDA achieves much better performance than Instant-NeuS, which could be faithful support for our analysis that we may need more flexible representation to model small structures that are with complicated topologies. We share some qualitative results in Figure.~\ref{fig:sota-cmp}. NeuDA is promising to capture fine-grained surface details. Especially, we can see that NeuDA successfully preserves the hollow structures of the ``BMVS Jade" object, while NeuS fills most of the holes with incorrect meshes. In Table~\ref{tab:psnr}, we present the case-by-case PSNR values following VolSDF \cite{yariv2021volume} as a supplementary experiment, where NeuDA yields a slightly lower mean PSNR than Instant-NeuS. We argue that capturing better renderings is out of the scope of this paper.

\subsection{Ablation Studies}
\label{ablation}

In this section, we ablate several major components of NeuDA under the w/ mask setting on the DTU dataset. 
\newline


\noindent \textbf{Deformable Anchors.} Table.~\ref{tab:ablation-da} discusses the effectiveness of our core contribution, \emph{i.e.}, deformable anchors, for surface reconstruction. ``Row 2 \emph{vs.} Row 3" indicates that storing the 3D position (or anchor) to each vertex of the voxel grid achieves similar performance to saving the feature vector (or embedding). In ``Row 4", we optimize the anchor points in ``Row 3" to secure the deformbale anchor representation. Both ``Row 3 \emph{vs.} Row 4" and ``Row 2 \emph{vs.} Row 4" shows that NeuDA is a more flexible scene representation method for implicit surface reconstruction.

\begin{figure}[t!]
  \centering
   \includegraphics[width=0.86\linewidth]{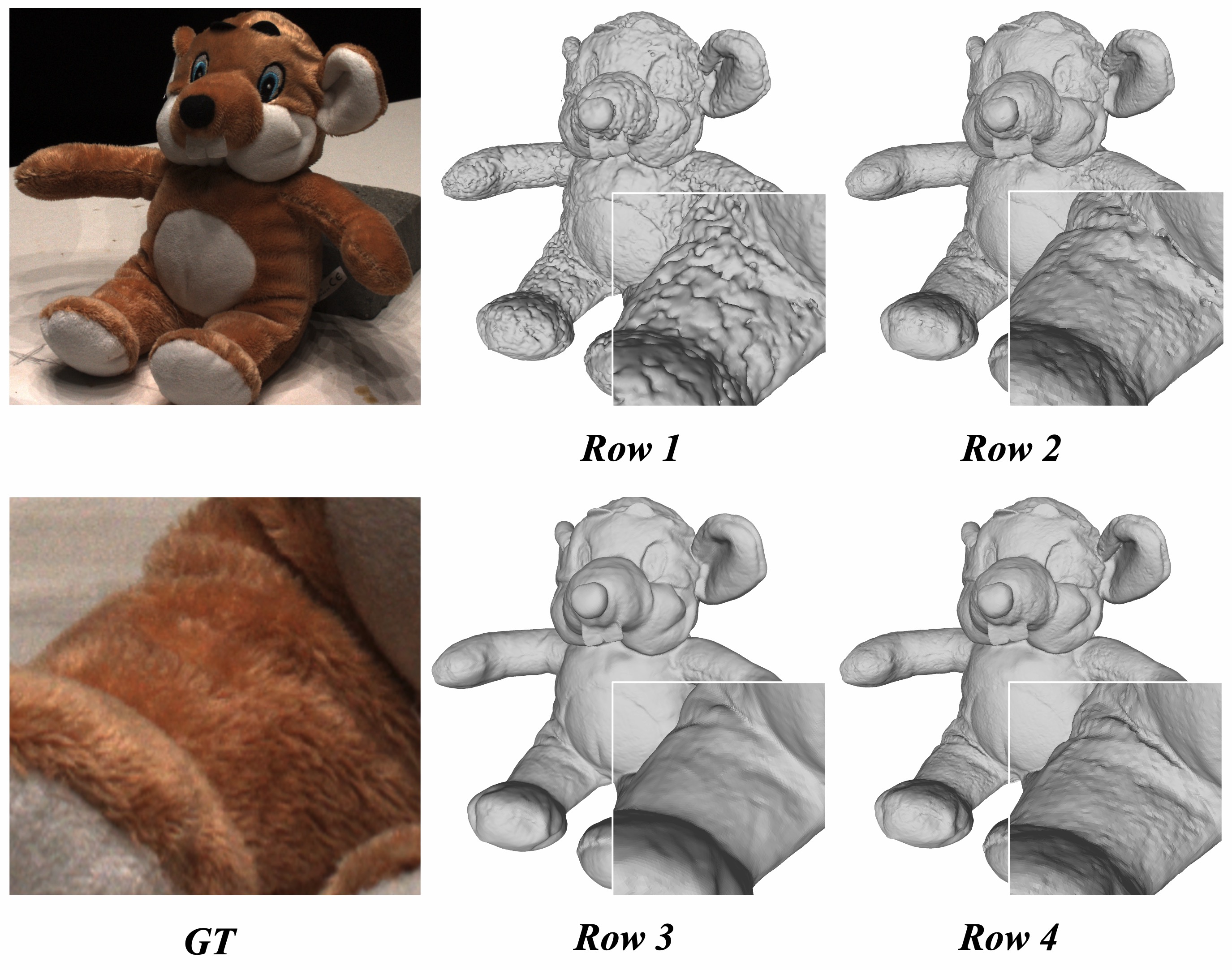}
   \caption{The qualitative explanation to ablations in Table~\ref{tab:ablation-da}.}
   \label{fig:da-cmp}
\end{figure}

\setlength\tabcolsep{5pt}
\begin{table}[t!]
\centering
\begin{tabular}{ c c c | c }
\toprule
Multi-Level & Opt. Grid & Feat. or Anc. & Mean CD \\
\midrule\midrule
& $\surd$ & Feat. & 1.25 \\
$\surd$ & $\surd$ & Feat. & 0.74 \\
$\surd$ & & Anc. & 0.75 \\
$\surd$ & $\surd$ & Anc. & 0.67 \\
\bottomrule
\end{tabular}

\caption{We study the core ``deformable anchors" technique (Row 4) by taking the standard multi-level feature grid method (Row 2) as the main baseline. ``Feats. or Anc." indicates that we store a feature vector or an anchor point to each vertex. ``Opt. Grid" means whether to optimize the maintained voxel grid or not. }
\label{tab:ablation-da}
\vspace{-0.3cm}
\end{table}

\begin{figure}[t]
  \centering
   \includegraphics[width=1.0\linewidth]{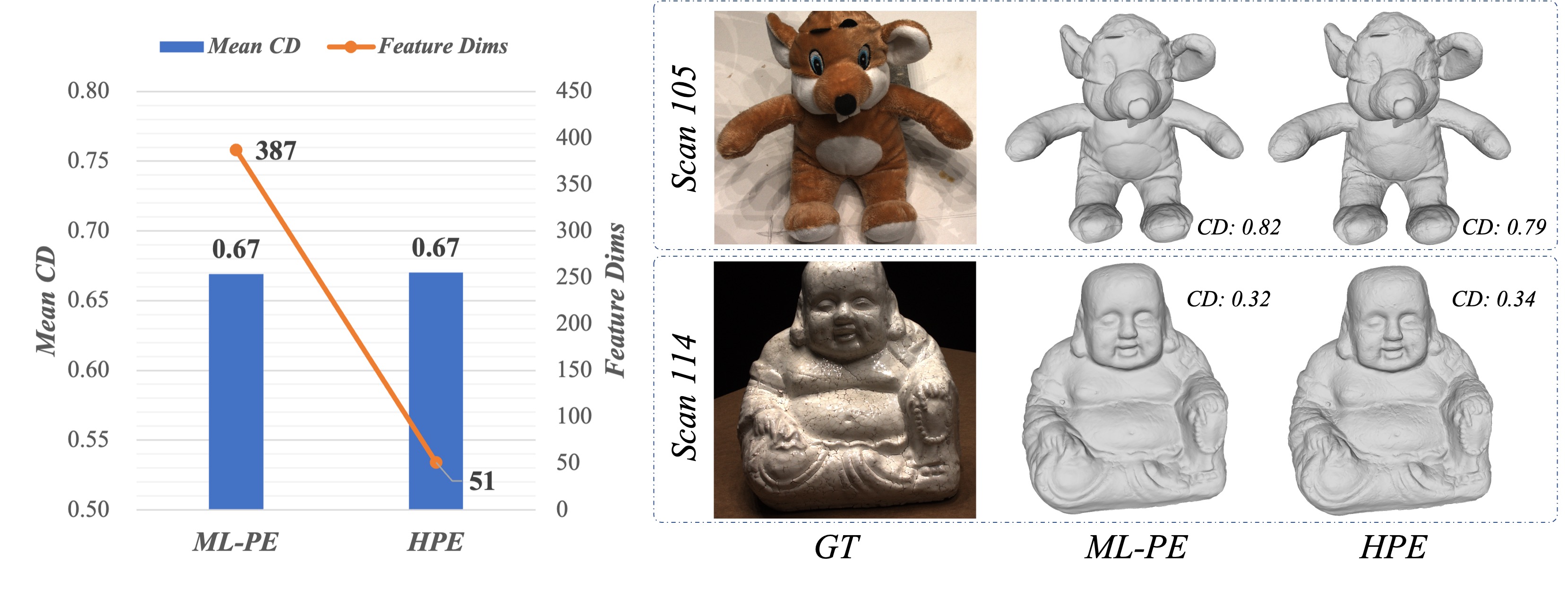}
   \caption{\textbf{ML-PE or HPE?} ML-PE means we employ the standard positional encoding function \cite{mildenhall2020nerf} instead of Eqn.~\ref{eqn:HPE-1} for NeuDA. HPE performs equally to ML-PE, while returns a low-dimension embedding vector.}
   \label{fig:ablation-hpe}
   \vspace{-0.3cm}
\end{figure}

\noindent \textbf{Hierarchical Position Encoding.} Figure.~\ref{fig:ablation-hpe} compares the lightweight HPE strategy and the standard multi-level positional encoding (ML-PE) approach. We find HPE and ML-PE perform equally on the DTU dataset. Both obtain a mean CD score of 0.67 and produce similar topologies. As analyzed before, it is possible that the standard encoding function \cite{mildenhall2020nerf} may contain some redundant information in the multi-level (or hierarchical) grid structure. Overall, HPE is sufficient to represent high-frequency variation in geometry.

\begin{figure}[t]
  \centering
   \includegraphics[width=0.95\linewidth]{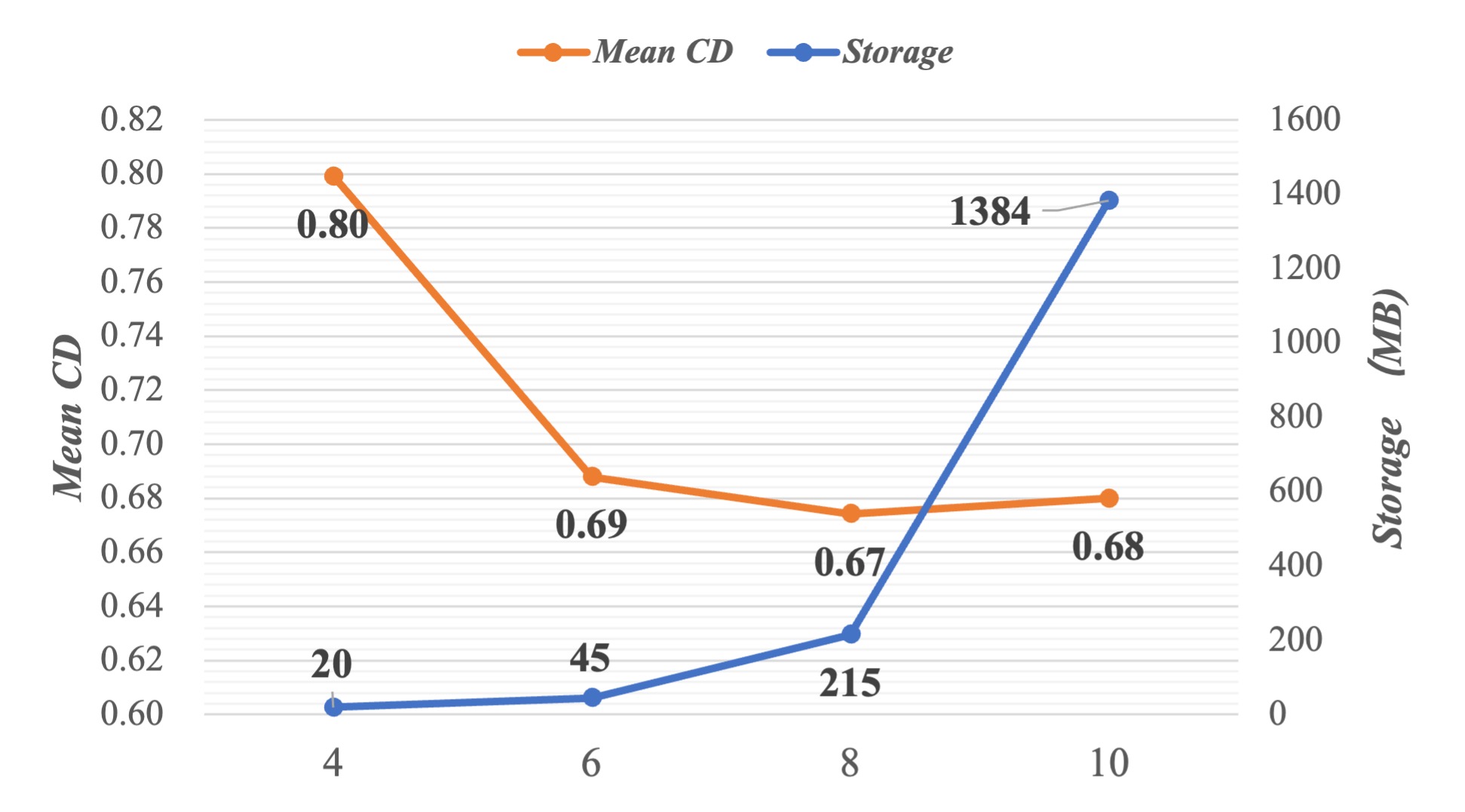}

   \caption{We report the storage cost and surface reconstruction quality w.r.t. the total grid levels.}
   \label{fig:ablation-level}
   \vspace{-0.3cm}
\end{figure}

\noindent \textbf{How many levels?} Figure~\ref{fig:ablation-level} explores the impact on surface reconstruction quality of different hierarchical levels $L$ for NeuDA. $L$ is a trade-off hyper-parameter for model size and capability. The performance increases with a higher level at first. However, when setting $L$ to 10, NeuDA produces a slightly lower Chamfer Distance. A possible reason is that there is much more redundant information in NeuDA-10, which might be harmful to the SDF approximation. 


\begin{figure}[h]
    \centering
    \includegraphics[width=0.47\textwidth]{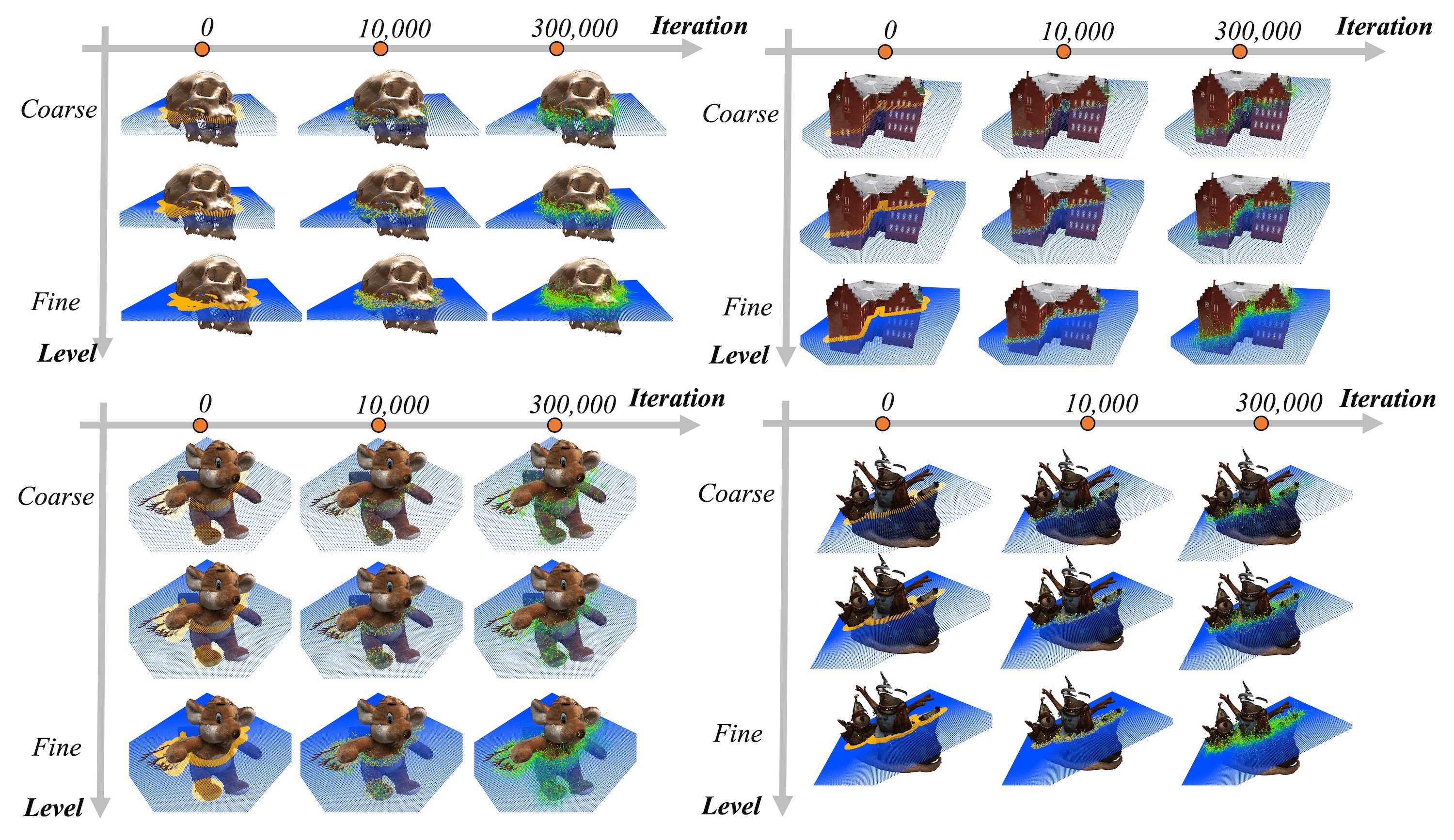}
    \vspace{-2mm}
    \caption{{Deformation Process of Anchor Points. The anchor points (e.g. orange points) are uniformly distributed in the 3D box at beginning and would move to object surfaces as training convergences. Zoom in for better view.  
    }}
    \label{fig:multi-level-DA}
    \vspace{-6mm}
\end{figure}

\begin{figure*}[t]
  \centering
   \includegraphics[width=1.0\linewidth]{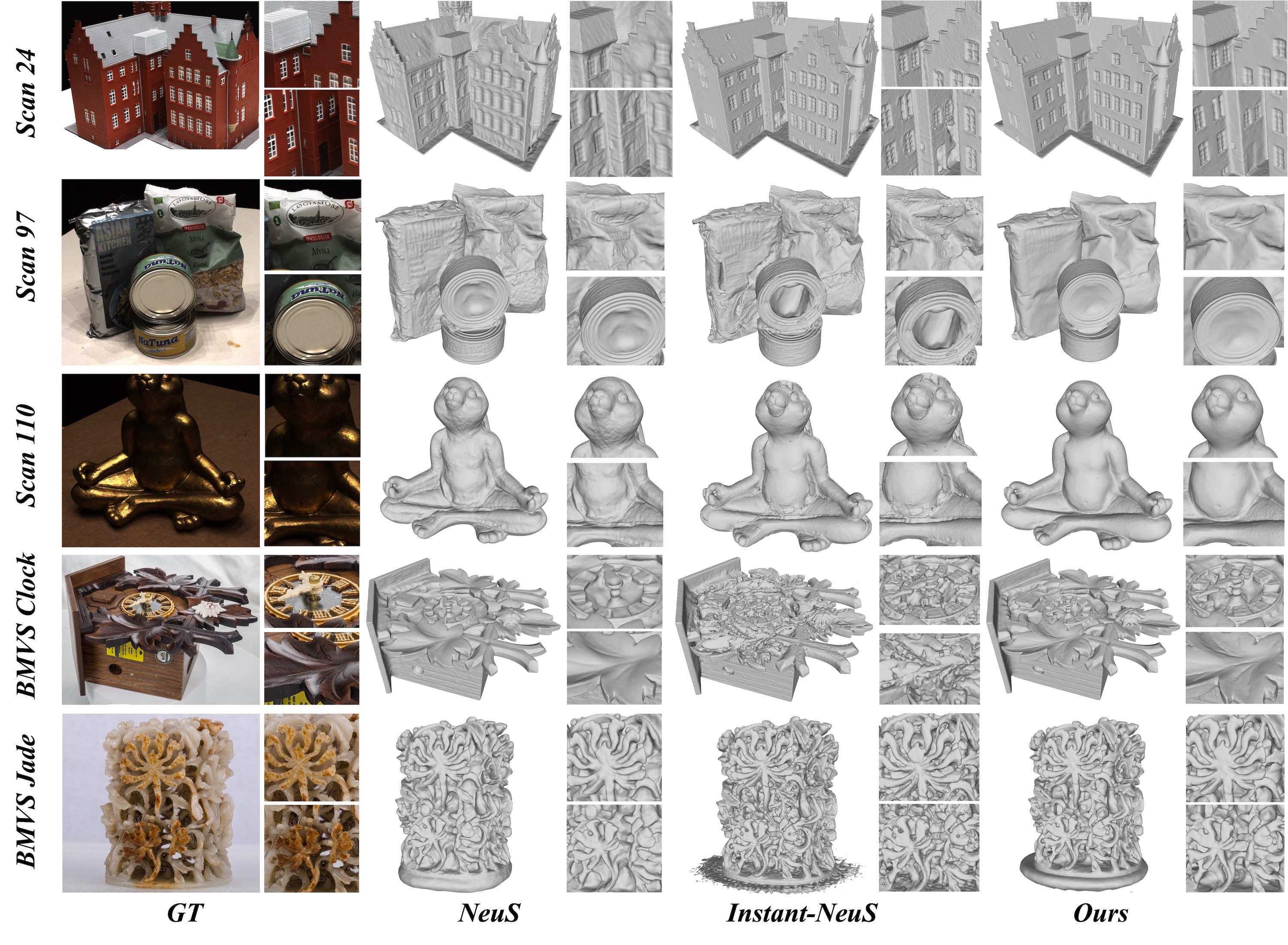}
   \caption{\textbf{Qualitative Comparisons on DTU and BlendedMVS (BMVS).} NeuDA can produce promising typologies for all the studied cases. Especially, it shows potential in handling fine-grained geometry structures such as small holes, curves, and other abrupt surface areas. Zoom in for a better view.}
   \label{fig:sota-cmp}
\end{figure*}

\section{Discussion \& Limitation}
One of the major limitations of this paper is that we follow an intuitive idea to propose NeuDA and conduct empirical studies to validate its performance. Although we can not provide strictly mathematical proof, we prudently respond to this concern and provide qualitative proof by reporting the anchor points' deformation process in Figure \ref{fig:multi-level-DA}. 

Taking a slice of grid voxels as an example, we can see the anchor points (\emph{e.g.} orange points) move to object surfaces as training convergences, resulting in an implied adaptive representation. 
Intuitively, the SDF change has an increasing effect on geometry prediction as the anchor approaches the surfaces, while the SDF change of a position far from the object has weak effects. Thus, the optimization process may force those anchors (``yellow" points) to move to positions nearly around the object surfaces to better reflect the SDF changes. 
The deformable anchor shares some similar concepts with deformable convolution \cite{dai2017deformable} and makes its movement process like a mesh deformation process.
Moreover, as each query point has eight anchors, from another perspective, each anchor follows an individual mesh deformation process. Thereby, NeuDA may play an important role in learning and ensembling multiple 3D reconstruction models. 



\section{Conclusion}
This paper studies neural implicit surface reconstruction. We find that previous works (\emph{e.g.} NeuS) are likely to produce over-smoothing surfaces for small local geometry structures and surface abrupt regions.
A possible reason is that the spatial context in 3D space has not been flexibly exploited. We take inspiration from the insight and propose NeuDA, namely \emph{Neural Deformable Anchors}, as a solution. NeuDA is leveraging multi-level voxel grids, and is empowered by the core ``Deformable Anchors (DA)" representation approach and a simple hierarchical position encoding strategy. The former maintains learnable anchor points at verities to enhance the capability of neural implicit model in handling complicated geometric structures, and the latter explores complementaries of high-frequency and low-frequency geometry properties in the multi-level anchor grid structure. The comparisons with baselines and SOTA methods demonstrate the superiority of NeuDA in capturing high-fidelity typologies.

{\small


\begin{thebibliography}{10}\itemsep=-1pt

\bibitem{barron2021mip}
Jonathan~T Barron, Ben Mildenhall, Matthew Tancik, Peter Hedman, Ricardo
  Martin-Brualla, and Pratul~P Srinivasan.
\newblock Mip-nerf: A multiscale representation for anti-aliasing neural
  radiance fields.
\newblock In {\em Proceedings of the IEEE/CVF International Conference on
  Computer Vision}, pages 5855--5864, 2021.

\bibitem{boulch2022poco}
Alexandre Boulch and Renaud Marlet.
\newblock Poco: Point convolution for surface reconstruction.
\newblock In {\em Proceedings of the IEEE/CVF Conference on Computer Vision and
  Pattern Recognition}, pages 6302--6314, 2022.

\bibitem{chen2022tensorf}
Anpei Chen, Zexiang Xu, Andreas Geiger, Jingyi Yu, and Hao Su.
\newblock Tensorf: Tensorial radiance fields.
\newblock In {\em Computer Vision--ECCV 2022: 17th European Conference, Tel
  Aviv, Israel, October 23--27, 2022, Proceedings, Part XXXII}, pages 333--350.
  Springer, 2022.

\bibitem{dai2017deformable}
Jifeng Dai, Haozhi Qi, Yuwen Xiong, Yi Li, Guodong Zhang, Han Hu, and Yichen
  Wei.
\newblock Deformable convolutional networks.
\newblock In {\em Proceedings of the IEEE international conference on computer
  vision}, pages 764--773, 2017.

\bibitem{darmon2022improving}
Fran{\c{c}}ois Darmon, B{\'e}n{\'e}dicte Bascle, Jean-Cl{\'e}ment Devaux,
  Pascal Monasse, and Mathieu Aubry.
\newblock Improving neural implicit surfaces geometry with patch warping.
\newblock In {\em Proceedings of the IEEE/CVF Conference on Computer Vision and
  Pattern Recognition}, pages 6260--6269, 2022.

\bibitem{fu_geo-neus_2022}
Qiancheng Fu, Qingshan Xu, Yew-Soon Ong, and Wenbing Tao.
\newblock Geo-{Neus}: {Geometry}-{Consistent} {Neural} {Implicit} {Surfaces}
  {Learning} for {Multi}-view {Reconstruction}, May 2022.
\newblock Number: arXiv:2205.15848 arXiv:2205.15848 [cs].

\bibitem{gropp_implicit_2020}
Amos Gropp, Lior Yariv, Niv Haim, Matan Atzmon, and Yaron Lipman.
\newblock Implicit {Geometric} {Regularization} for {Learning} {Shapes}.
\newblock {\em arXiv:2002.10099 [cs, stat]}, July 2020.
\newblock arXiv: 2002.10099.

\bibitem{jensen_dtu}
Rasmus Jensen, Anders Dahl, George Vogiatzis, Engil Tola, and Henrik Aanæs.
\newblock Large scale multi-view stereopsis evaluation.
\newblock In {\em 2014 IEEE Conference on Computer Vision and Pattern
  Recognition}, pages 406--413, 2014.

\bibitem{kazhdan2006poisson}
Michael Kazhdan, Matthew Bolitho, and Hugues Hoppe.
\newblock Poisson surface reconstruction.
\newblock In {\em Proceedings of the fourth Eurographics symposium on Geometry
  processing}, volume~7, 2006.

\bibitem{kellnhofer2021neural}
Petr Kellnhofer, Lars~C Jebe, Andrew Jones, Ryan Spicer, Kari Pulli, and Gordon
  Wetzstein.
\newblock Neural lumigraph rendering.
\newblock In {\em Proceedings of the IEEE/CVF Conference on Computer Vision and
  Pattern Recognition}, pages 4287--4297, 2021.

\bibitem{kingma2014adam}
Diederik~P Kingma and Jimmy Ba.
\newblock Adam: A method for stochastic optimization.
\newblock {\em arXiv preprint arXiv:1412.6980}, 2014.

\bibitem{liang_spidr_2022}
Ruofan Liang, Jiahao Zhang, Haoda Li, Chen Yang, and Nandita Vijaykumar.
\newblock {SPIDR}: {SDF}-based {Neural} {Point} {Fields} for {Illumination} and
  {Deformation}, Oct. 2022.
\newblock arXiv:2210.08398 [cs].

\bibitem{ling2022shadowneus}
Jingwang Ling, Zhibo Wang, and Feng Xu.
\newblock Shadowneus: Neural sdf reconstruction by shadow ray supervision.
\newblock {\em arXiv preprint arXiv:2211.14086}, 2022.

\bibitem{liu2020neural}
Lingjie Liu, Jiatao Gu, Kyaw Zaw~Lin, Tat-Seng Chua, and Christian Theobalt.
\newblock Neural sparse voxel fields.
\newblock {\em Advances in Neural Information Processing Systems},
  33:15651--15663, 2020.

\bibitem{long2022neuraludf}
Xiaoxiao Long, Cheng Lin, Lingjie Liu, Yuan Liu, Peng Wang, Christian Theobalt,
  Taku Komura, and Wenping Wang.
\newblock Neuraludf: Learning unsigned distance fields for multi-view
  reconstruction of surfaces with arbitrary topologies.
\newblock {\em arXiv preprint arXiv:2211.14173}, 2022.

\bibitem{mescheder_occupancy_2019}
Lars Mescheder, Michael Oechsle, Michael Niemeyer, Sebastian Nowozin, and
  Andreas Geiger.
\newblock Occupancy {Networks}: {Learning} {3D} {Reconstruction} in {Function}
  {Space}.
\newblock In {\em 2019 {IEEE}/{CVF} {Conference} on {Computer} {Vision} and
  {Pattern} {Recognition} ({CVPR})}, pages 4455--4465, Long Beach, CA, USA,
  June 2019. IEEE.

\bibitem{mildenhall2020nerf}
Ben Mildenhall, Pratul~P Srinivasan, Matthew Tancik, Jonathan~T Barron, Ravi
  Ramamoorthi, and Ren Ng.
\newblock Nerf: Representing scenes as neural radiance fields for view
  synthesis.
\newblock In {\em ECCV}, pages 405--421. Springer, 2020.

\bibitem{mueller2022instant}
Thomas M\"uller, Alex Evans, Christoph Schied, and Alexander Keller.
\newblock Instant neural graphics primitives with a multiresolution hash
  encoding.
\newblock {\em ACM Trans. Graph.}, 41(4):102:1--102:15, July 2022.

\bibitem{niemeyer_differentiable_2020}
Michael Niemeyer, Lars Mescheder, Michael Oechsle, and Andreas Geiger.
\newblock Differentiable {Volumetric} {Rendering}: {Learning} {Implicit} {3D}
  {Representations} {Without} {3D} {Supervision}.
\newblock In {\em 2020 {IEEE}/{CVF} {Conference} on {Computer} {Vision} and
  {Pattern} {Recognition} ({CVPR})}, pages 3501--3512, Seattle, WA, USA, June
  2020. IEEE.

\bibitem{oechsle2021unisurf}
Michael Oechsle, Songyou Peng, and Andreas Geiger.
\newblock Unisurf: Unifying neural implicit surfaces and radiance fields for
  multi-view reconstruction.
\newblock In {\em Proceedings of the IEEE/CVF International Conference on
  Computer Vision}, pages 5589--5599, 2021.

\bibitem{park_deepsdf_2019}
Jeong~Joon Park, Peter Florence, Julian Straub, Richard Newcombe, and Steven
  Lovegrove.
\newblock {DeepSDF}: {Learning} {Continuous} {Signed} {Distance} {Functions}
  for {Shape} {Representation}.
\newblock In {\em 2019 {IEEE}/{CVF} {Conference} on {Computer} {Vision} and
  {Pattern} {Recognition} ({CVPR})}, pages 165--174, Long Beach, CA, USA, June
  2019. IEEE.

\bibitem{rosu2022hashsdf}
Radu~Alexandru Rosu and Sven Behnke.
\newblock Hashsdf: Accurate implicit surfaces with fast local features on
  permutohedral lattices.
\newblock {\em arXiv preprint arXiv:2211.12562}, 2022.

\bibitem{yu_and_fridovichkeil2021plenoxels}
{Sara Fridovich-Keil and Alex Yu}, Matthew Tancik, Qinhong Chen, Benjamin
  Recht, and Angjoo Kanazawa.
\newblock Plenoxels: Radiance fields without neural networks.
\newblock In {\em CVPR}, 2022.

\bibitem{schonberger_structure--motion_2016}
Johannes~L. Schonberger and Jan-Michael Frahm.
\newblock Structure-from-{Motion} {Revisited}.
\newblock In {\em 2016 {IEEE} {Conference} on {Computer} {Vision} and {Pattern}
  {Recognition} ({CVPR})}, pages 4104--4113, Las Vegas, NV, USA, June 2016.
  IEEE.

\bibitem{sun2022direct}
Cheng Sun, Min Sun, and Hwann-Tzong Chen.
\newblock Direct voxel grid optimization: Super-fast convergence for radiance
  fields reconstruction.
\newblock In {\em Proceedings of the IEEE/CVF Conference on Computer Vision and
  Pattern Recognition}, pages 5459--5469, 2022.

\bibitem{takikawa2021neural}
Towaki Takikawa, Joey Litalien, Kangxue Yin, Karsten Kreis, Charles Loop, Derek
  Nowrouzezahrai, Alec Jacobson, Morgan McGuire, and Sanja Fidler.
\newblock Neural geometric level of detail: Real-time rendering with implicit
  3d shapes.
\newblock In {\em Proceedings of the IEEE/CVF Conference on Computer Vision and
  Pattern Recognition}, pages 11358--11367, 2021.

\bibitem{verbin2022ref}
Dor Verbin, Peter Hedman, Ben Mildenhall, Todd Zickler, Jonathan~T Barron, and
  Pratul~P Srinivasan.
\newblock Ref-nerf: Structured view-dependent appearance for neural radiance
  fields.
\newblock In {\em 2022 IEEE/CVF Conference on Computer Vision and Pattern
  Recognition (CVPR)}, pages 5481--5490. IEEE, 2022.

\bibitem{wang2021neus}
Peng Wang, Lingjie Liu, Yuan Liu, Christian Theobalt, Taku Komura, and Wenping
  Wang.
\newblock Neus: Learning neural implicit surfaces by volume rendering for
  multi-view reconstruction.
\newblock {\em NeurIPS}, 2021.

\bibitem{wang2022neus2}
Yiming Wang, Qin Han, Marc Habermann, Kostas Daniilidis, Christian Theobalt,
  and Lingjie Liu.
\newblock Neus2: Fast learning of neural implicit surfaces for multi-view
  reconstruction.
\newblock {\em arXiv preprint arXiv:2212.05231}, 2022.

\bibitem{wang_hf-neus_2022}
Yiqun Wang, Ivan Skorokhodov, and Peter Wonka.
\newblock {HF}-{NeuS}: {Improved} {Surface} {Reconstruction} {Using}
  {High}-{Frequency} {Details}, Sept. 2022.
\newblock arXiv:2206.07850 [cs].

\bibitem{wu_voxurf_2022}
Tong Wu, Jiaqi Wang, Xingang Pan, Xudong Xu, Christian Theobalt, Ziwei Liu, and
  Dahua Lin.
\newblock Voxurf: {Voxel}-based {Efficient} and {Accurate} {Neural} {Surface}
  {Reconstruction}, Aug. 2022.
\newblock arXiv:2208.12697 [cs].

\bibitem{xu2022point}
Qiangeng Xu, Zexiang Xu, Julien Philip, Sai Bi, Zhixin Shu, Kalyan Sunkavalli,
  and Ulrich Neumann.
\newblock Point-nerf: Point-based neural radiance fields.
\newblock In {\em Proceedings of the IEEE/CVF Conference on Computer Vision and
  Pattern Recognition}, pages 5438--5448, 2022.

\bibitem{yao2020blendedmvs}
Yao Yao, Zixin Luo, Shiwei Li, Jingyang Zhang, Yufan Ren, Lei Zhou, Tian Fang,
  and Long Quan.
\newblock Blendedmvs: A large-scale dataset for generalized multi-view stereo
  networks.
\newblock In {\em Proceedings of the IEEE/CVF Conference on Computer Vision and
  Pattern Recognition}, pages 1790--1799, 2020.

\bibitem{yariv2021volume}
Lior Yariv, Jiatao Gu, Yoni Kasten, and Yaron Lipman.
\newblock Volume rendering of neural implicit surfaces.
\newblock In {\em Thirty-Fifth Conference on Neural Information Processing
  Systems}, 2021.

\bibitem{yariv2020multiview}
Lior Yariv, Yoni Kasten, Dror Moran, Meirav Galun, Matan Atzmon, Basri Ronen,
  and Yaron Lipman.
\newblock Multiview neural surface reconstruction by disentangling geometry and
  appearance.
\newblock {\em Advances in Neural Information Processing Systems}, 33, 2020.

\bibitem{yu2021plenoctrees}
Alex Yu, Ruilong Li, Matthew Tancik, Hao Li, Ren Ng, and Angjoo Kanazawa.
\newblock {PlenOctrees} for real-time rendering of neural radiance fields.
\newblock In {\em ICCV}, 2021.

\bibitem{Yu2022MonoSDF}
Zehao Yu, Songyou Peng, Michael Niemeyer, Torsten Sattler, and Andreas Geiger.
\newblock Monosdf: Exploring monocular geometric cues for neural implicit
  surface reconstruction.
\newblock {\em Advances in Neural Information Processing Systems (NeurIPS)},
  2022.

\bibitem{zhang_learning_2021}
Jingyang Zhang, Yao Yao, and Long Quan.
\newblock Learning {Signed} {Distance} {Field} for {Multi}-view {Surface}
  {Reconstruction}.
\newblock In {\em 2021 {IEEE}/{CVF} {International} {Conference} on {Computer}
  {Vision} ({ICCV})}, pages 6505--6514, Montreal, QC, Canada, Oct. 2021. IEEE.

\end{thebibliography}
}


\pagebreak


\beginsupplement


\twocolumn[{%
\renewcommand\twocolumn[1][]{#1}%
\maketitle
\begin{center}
    \centering
    \includegraphics[width=\textwidth]{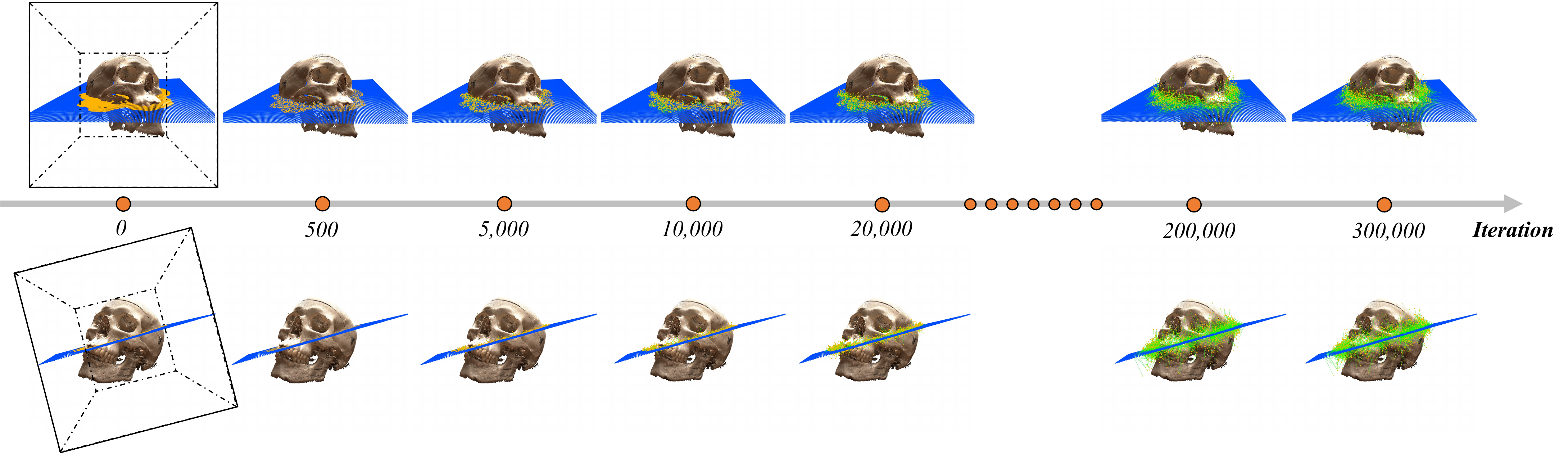}\vspace{-0.3cm}
    \captionof{figure}{\textbf{Deformation Process of Anchor Points (Scan 65).} The anchor points (\emph{e.g.} orange points) are uniformly distributed in the 3D box at beginning, and would move to object surfaces as training convergences. The observation could be a faithful support to our deformable anchor analyzes. Zoom in for a better view.}
    \label{fig:da_movements}
\end{center}%
}]


\section{Qualitative Proof for Deformable Anchors}
\label{sec:da_proof}
As discussed in Sec. {\color{red} 4.3} in the main paper, deformable anchors enable NeuDA to become a more flexible scene representation approach for surface reconstruction. Here, we provide qualitative proof by reporting the anchor points' deformation process in S.Figure \ref{fig:da_movements}. Taking a slice of grid voxels as an example, we can see the anchor points (\emph{e.g.} orange points) are uniformly distributed in the 3D box at beginning, and would move to object surfaces as training convergences. This observation should be faithful support that deformable anchors optimized through backpropagation can adaptively represent surface geometries and achieve more flexibility in modeling fine-grained geometric structures.

\section{Discussion: Standard Deviation}

\begin{figure}[h]
  \centering
   \includegraphics[width=1.0\linewidth]{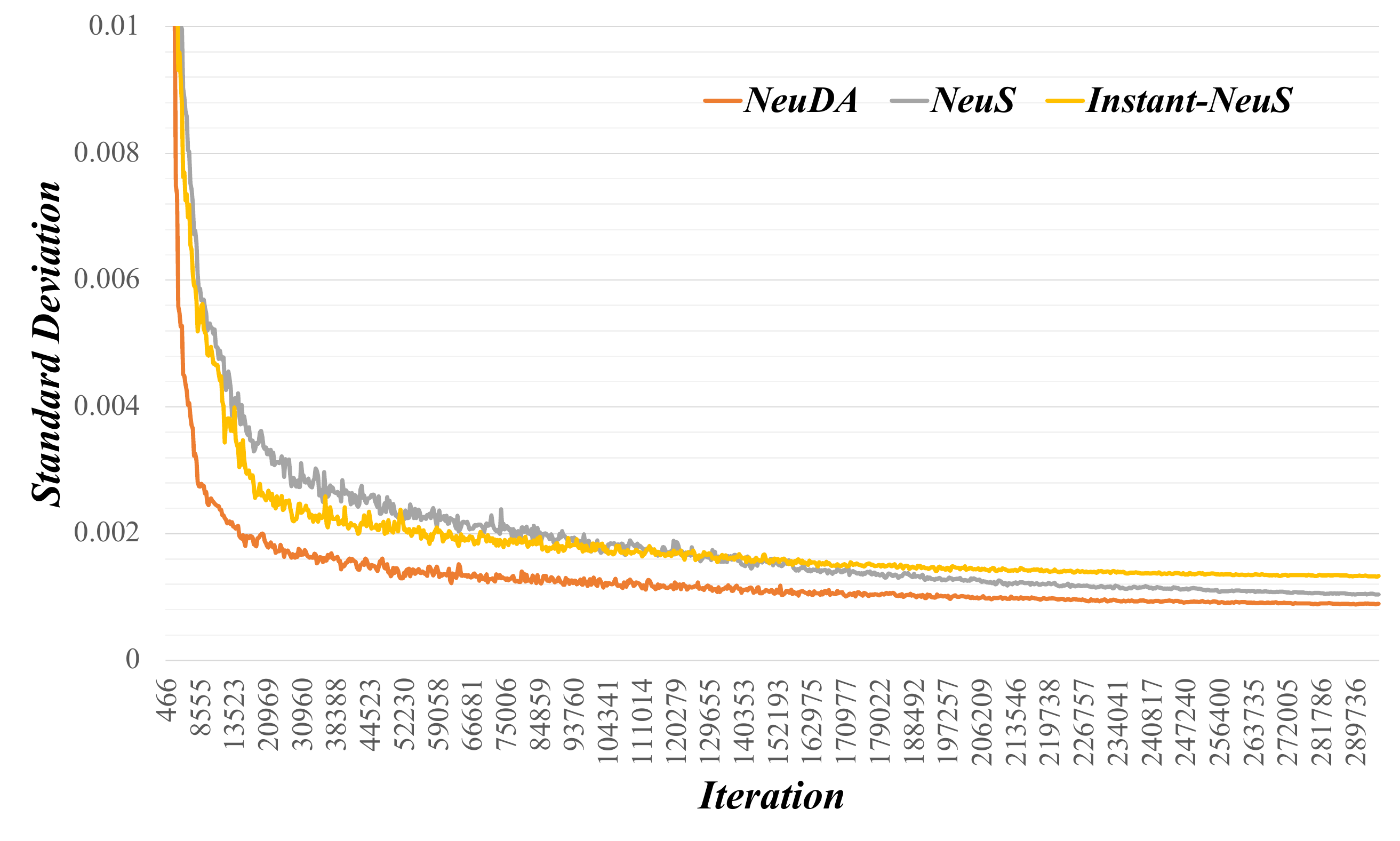}

   \caption{\textbf{Discussion: Standard Deviation (Scan 65).}
   The standard deviation of NeuDA converges rapidly and achieves the lowest value compared to NeuS \cite{wang2021neus} and Instant-NeuS \cite{mueller2022instant,wang2021neus}. NeuS \cite{wang2021neus} indicates that lower standard deviation means more clear and sharper surface. }
   \label{fig:stand_deviation}
   \vspace{-0.3cm}
\end{figure}

Following NeuS \cite{wang2021neus} (See Sec. \textcolor{red}{E.4} and Figure \textcolor{red}{14} in their paper), we report the curves of standard deviation for different methods in S.Figure \ref{fig:stand_deviation} to evaluate the sharpness of the reconstructed surface. NeuDA converges rapidly and yields the lowest value compared to NeuS and Instant-NeuS, which means NeuDA can produce more clear and sharper surfaces with less time cost. We find the ``standard deviation curve" might not directly reflect to the global reconstruction quality, as Instant-NeuS yields a slightly better mean CD score than NeuS.
\newline

\noindent \textbf{Quoted Texts from NeuS:} \emph{As we can see, the optimization process will automatically reduce the standard
deviation so that the surface becomes more clear and sharper with more training steps.}

\section{Implementation Details}
\label{sec:implement}

NeuDA consists of the hierarchical ``deformable anchors" representation, an SDF network, and a color network. The hierarchical deformable anchors are arranged into \emph{L} levels (\emph{L} set to 8 as default), each contains \emph{T} coordinate vectors, \emph{e.g.}, $(x,y,z)$. The number of coordinate vectors \emph{T} is set to 16 at the coarsest level and is growing by $1.38\times$ than its coarser level. The deformable anchors and view directions are encoded by positional encoding with 8 frequencies and 4 frequencies, respectively. The signed distance function is approximated by the 4-layer MLPs with hidden-layer size of 256. We additionally predict a normal vector from the SDF network, and using it to construct a normal regularization loss defined in Eqn. {\color{red}11} in the main paper. The color network follows a similar architecture as NeuS \cite{wang2021neus}, including 4 layers with size of 256. We adopt the Adam optimizer \cite{kingma2014adam} to opmize the model. We train NeuDA in 300k iterations and decay the learning rate from $5\times 10^{-4}$ to $2.5\times 10^{-5}$ via the cosine decay scheduler.


\section{More Qualitative Results}
\begin{figure*}[thb!]
  \centering
   \includegraphics[width=1.0\linewidth]{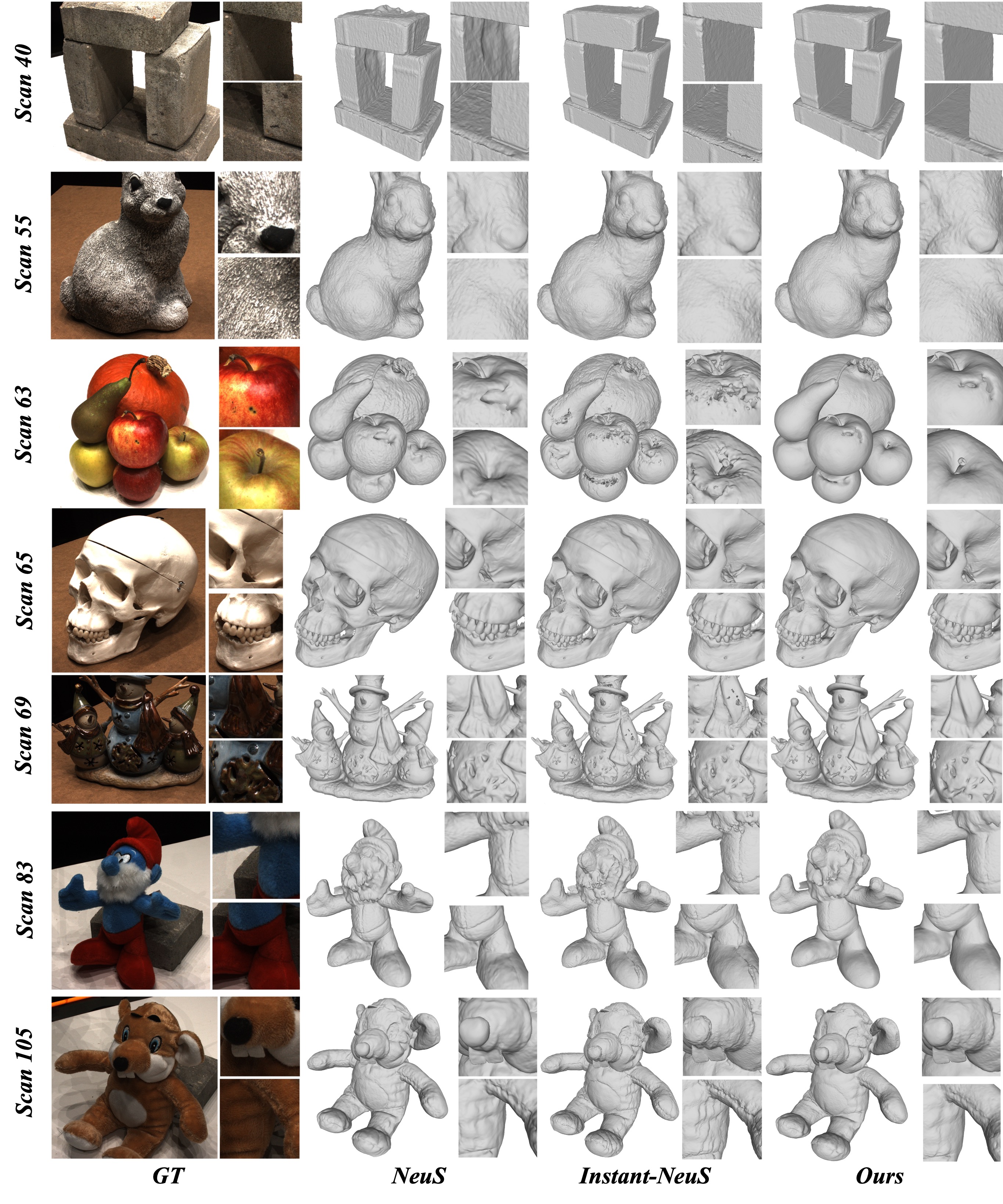}
   \caption{More surface reconstruction comparisons on DTU. (Part 1/2)}
   \label{fig:s-dtu-part1}
\end{figure*}

\begin{figure*}[thb!]
  \centering
   \includegraphics[width=1.0\linewidth]{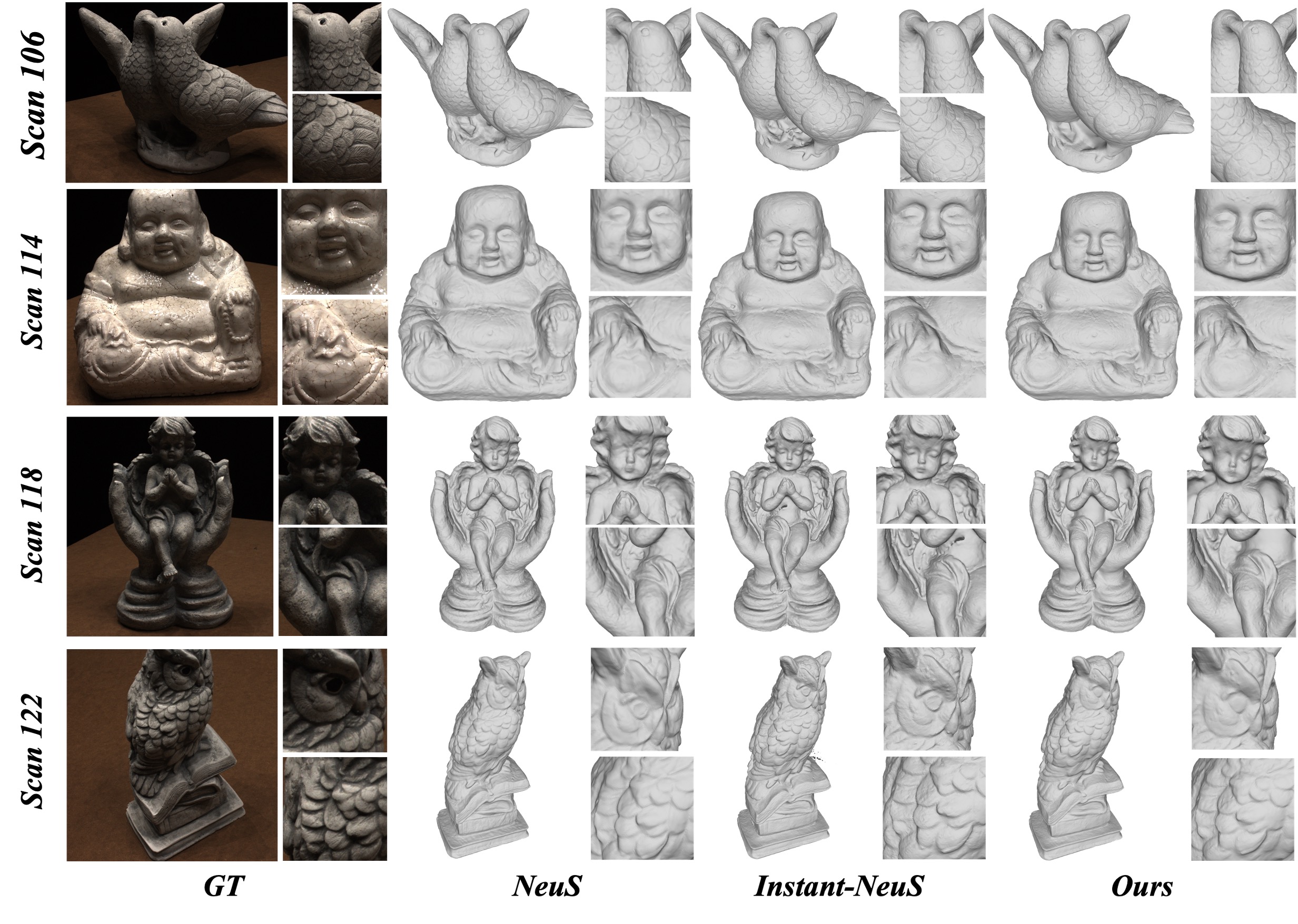}
   \caption{More surface reconstruction comparisons on DTU. (Part 2/2)}
   \label{fig:s-dtu-part2}
\end{figure*}

\begin{figure*}[thb!]
  \centering
   \includegraphics[width=1.0\linewidth]{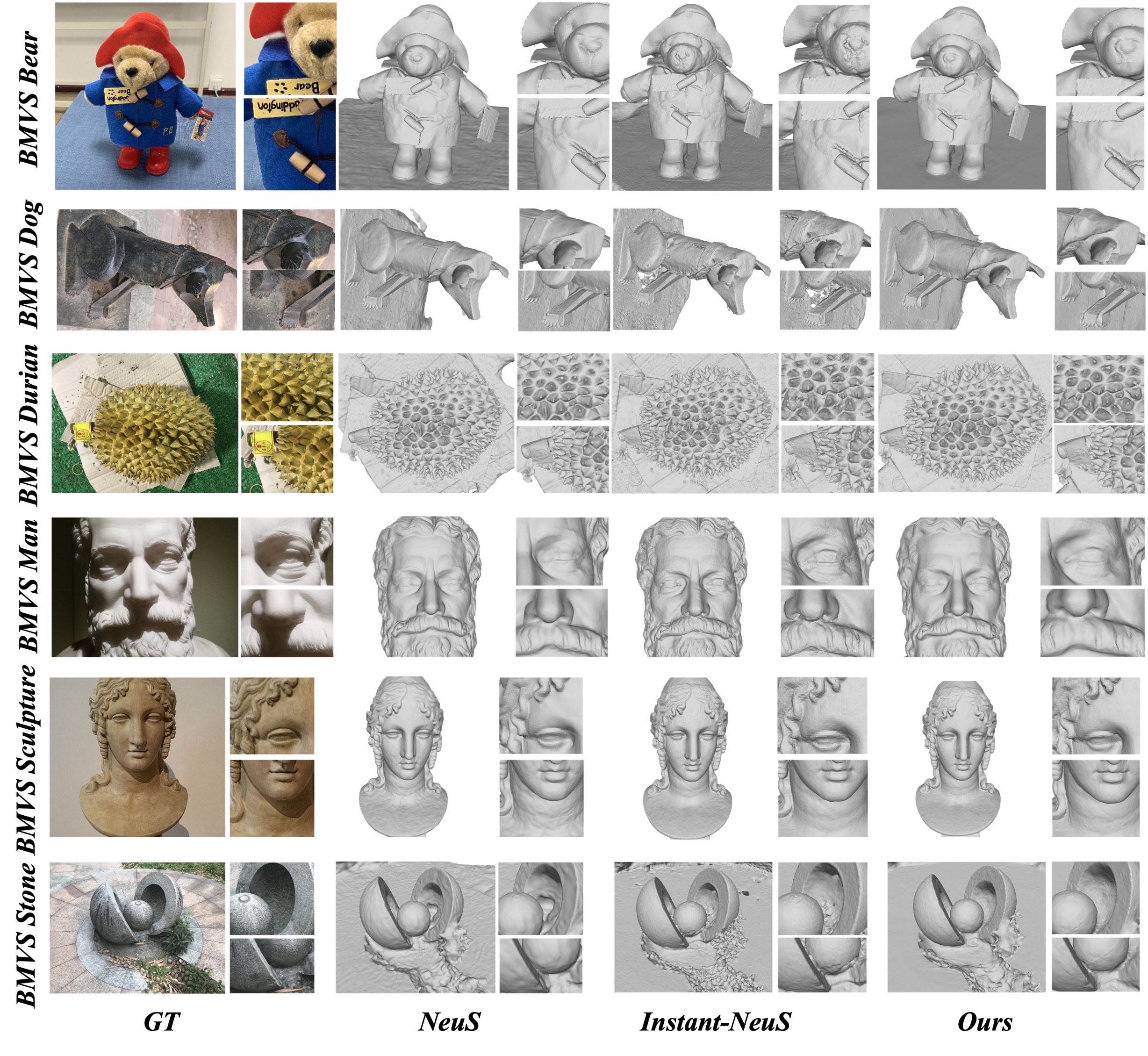}
   \caption{More surface reconstruction comparisons on BlendedMVS.}
   \label{fig:s-bmvs}
\end{figure*}

S.Figure \ref{fig:s-dtu-part1}, S.Figure \ref{fig:s-dtu-part2}, and S.Figure \ref{fig:s-bmvs} present the qualitative results of the remained cases on the DTU \cite{jensen_dtu} and BlendedMVS \cite{yao2020blendedmvs} datasets for \textcolor{red}{comprehensiveness}. Though NeuDA outperforms NeuS and Instant-NeuS by large margins quantitatively, the qualitative improvements for some cases are not really obvious.

\end{document}